\documentclass[
]{ceurart}

\usepackage[frozencache=true,cachedir=minted-cache]{minted} 
\usepackage{caption}
\usepackage{subcaption}
\usepackage{url}
\setminted{breaklines=true}

\begin{document}

\copyrightyear{2021}
\copyrightclause{Copyright for this paper by its authors.
  Use permitted under Creative Commons License Attribution 4.0
  International (CC BY 4.0).}

\conference{Forum for Information Retrieval Evaluation, December 13-17, 2021, India}

\title{Combining Textual Features for the Detection of Hateful and Offensive Language}

\author[1,2]{Sherzod Hakimov}[%
orcid=0000-0002-7421-6213,
email=sherzod.hakimov@tib.eu,
]
\address[1]{TIB - Leibniz Information Centre for Science and Technology, Hannover, Germany}
\address[2]{Leibniz University Hannover, L3S Research Center, Hannover, Germany}

\author[1,2]{Ralph Ewerth}[%
orcid=0000-0003-0918-6297,
email=ralph.ewerth@tib.eu,
]

\begin{abstract}
The detection of offensive, hateful and profane language has become a critical challenge since many users in social networks are exposed to cyberbullying activities on a daily basis. In this paper, we present an analysis of combining different textual features for the detection of hateful or offensive posts on Twitter. We provide a detailed experimental evaluation to understand the impact of each building block in a neural network architecture. The proposed architecture is evaluated on the \textit{English Subtask 1A: Identifying Hate, offensive and profane content from the post datasets} of \textit{HASOC-2021} dataset under the team name \textit{TIB-VA}. We compared different variants of the contextual word embeddings combined with the character level embeddings and the encoding of collected hate terms. 

\end{abstract}


\maketitle

\section{Introduction}

The detection of hateful, offensive and profane language has become a significant research challenge with the widespread usage of social media. Certain groups of people become targets of cyberbullying activities on a daily basis on many social networks such as Facebook, Twitter, or Instagram~\cite{kowalski2014bullying}. There have been many efforts by the research community and social media companies such as Facebook\footnote{\url{https://www.facebook.com/communitystandards/hate_speech}} and Twitter\footnote{\url{https://help.twitter.com/en/rules-and-policies/hateful-conduct-policy}} to the scope of hate speech and tackle the problem. In general, hate speech is defined as \textit{a language used to express hatred towards a targeted group or individuals based on specific characteristics such as religion, ethnicity, origin, sexual orientation, gender, physical appearance, disability or disease}~\cite{DBLP:conf/www/BadjatiyaG0V17,text_hs_1,DBLP:journals/csur/FortunaN18,warner2012detecting, DBLP:journals/lre/PolettoBSBP21, DBLP:journals/semweb/ZhangL19, DBLP:conf/acl/ZeinertID20}.

In this paper, we analyze the effects of combining multiple textual features to detect hateful, offensive or profane language expressed in the tweet text. We evaluated our approach on the \textit{Hate Speech and Offensive Content Identification in English and Indo-Aryan Languages} (HASOC) challenge datasets\footnote{\url{https://hasocfire.github.io}}. We submitted our solution to the \textit{English Subtask 1A: Identifying Hate, offensive and profane content from the post}~\cite{hasoc2021overview} of the HASOC-2021~\cite{hasoc2021} challenge series. The task involves classifying a given tweet text whether the content is hateful, offensive, or profane language or not. We proposed a combination of multiple textual features based on neural network architecture and evaluated different configurations. Our experimental evaluation is performed on all three datasets: HASOC-2019~\cite{hasoc2019}, HASOC-2020~\cite{hasoc2020}, HASOC-2021~\cite{hasoc2021}. 

The remainder of the paper is structured as follows. In Section~\ref{sec:approach}, we describe the proposed model architecture. In Section~\ref{sec:results}, the experimental setup, challenge datasets, as well as evaluations of model architectures are described in detail. Finally, the Section~\ref{sec:conclusion} concludes the paper.

\section{Approach}\label{sec:approach}

\begin{figure}[!ht]
	\centering
    \includegraphics[width=1.00\linewidth]{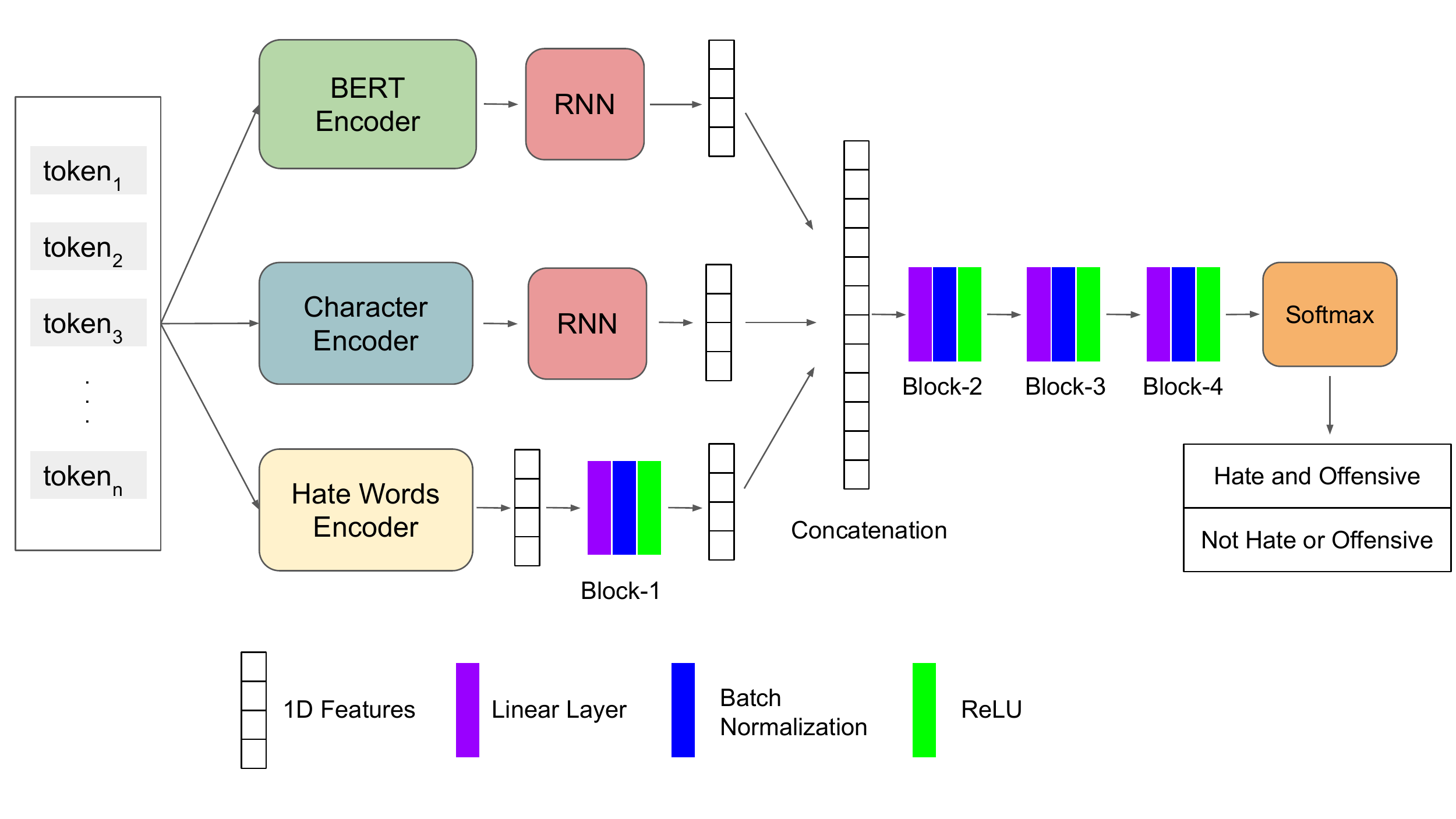}
	\caption{The model architecture combines character, hate words, and BERT embeddings that outputs probability of a given text being hate and offensive or not.}
	\label{fig:model}
\end{figure}

Our model architecture is built on top of three textual features that are combined to predict whether a given text contains hateful, offensive or profane language. The neural network architecture is shown in Figure~\ref{fig:model}. Input tokens are fed into \textit{BERT}, \textit{Character} and \textit{Hate Words} encoders to extract feature-specific vector representations. Once each feature representation is extracted, the outputs are fed into separate components to obtain one-dimensional vector representations. These vectors are concatenated and fed into three different blocks to obtain binary class probabilities. Each block is composed of a linear layer, batch normalization and a \textit{ReLU} activation function. The source code and the resources described below are shared publicly with the community\footnote{\url{https://github.com/sherzod-hakimov/HASOC-2021---Hate-Speech-Detection}}. Next, we describe the textual encoders in detail.

\noindent\textbf{BERT Encoder}: We used a pre-trained BERT~\cite{bert} model to obtain contextual 768-dimensional word vectors for each input token.\\
\noindent\textbf{Character Encoder}: Each input token is converted into vector representation based on the one-hot encoding of characters in English. We only use letters (a-z) to obtain a sequence of character-level vectors. \\
\noindent\textbf{Hate Words Encoder}: We collected a list of hate terms by combining the dictionary provided by \citet{multimodal_hs_2} with additional online dictionaries\footnote{\url{https://www.noswearing.com/dictionary} \& \url{https://hatebase.org/}}.
We manually filtered out terms that do not express hate concepts and obtained a list of \textit{1493} hate terms. The list contains a variety of terms with different lexical variations to increase the coverage of detecting such terms in tweets, e.g., \textit{bitc*}, \textit{border jumper}, \textit{nig**}, or \textit{chin*}. 
This encoder outputs a 1493-dimensional vector, a multi-hot encoding of hate terms in input tokens.

\section{Experimental Setup and Results}\label{sec:results}

In this section, we present the challenge datasets, details about data preprocessing and model parameters, and finally explain the experimental setup along with the obtained results.

\subsection{Datasets}

Our model architecture is built for the HASOC-2021~\cite{hasoc2021} \textit{English Subtask 1A: Identifying Hate, offensive and profane content from the post}~\cite{hasoc2021overview}. In Table~\ref{tab:dataset_stat}, we provide the number of data points for HASOC-2019\cite{hasoc2019}, HASOC-2020~\cite{hasoc2020}, and HASOC-2021~\cite{hasoc2021} editions of the challenge. The datasets include tweet text as input data and two target labels: \textit{Hate and Offensive (HOF)} and \textit{Not Hate or Offensive (NOT)}.

The number of training samples for the 2019 and 2021 editions are not equally distributed among the two classes. To overcome the class imbalance issue, we applied the oversampling method to the training splits. We randomly selected a certain number of data points for the minority class (HOF for 2019, NOT for 2021) and duplicated them to equalize with the number of data points for the majority class.

\begin{table}[!ht]
\caption{Distribution of data points for train and test splits for \textit{English Subtask 1A} for all editions of the HASOC datasets. \textbf{HOF}: Hate and Offensive, \textbf{NOT}: Not Hate or Offensive}
\begin{tabular}{|l|l|l|l|l|}
\hline
\multicolumn{1}{|c|}{\multirow{2}{*}{\textbf{Dataset}}} & \multicolumn{2}{c|}{\textbf{Train}} & \multicolumn{2}{c|}{\textbf{Test}} \\ \cline{2-5} 
\multicolumn{1}{|c|}{} & HOF & NOT & HOF & NOT \\ \hline
HASOC-2019~\cite{hasoc2019} & 2261 & 3591 & 288 & 865 \\ \hline
HASOC-2020~\cite{hasoc2020} & 1856 & 1852 & 807 & 785 \\ \hline
HASOC-2021~\cite{hasoc2021} & 2501 & 1342 & 765 & 456 \\ \hline
\end{tabular}
\label{tab:dataset_stat}
\end{table}

\subsection{Data Preprocessing}

There are several challenges with working text from Twitter. In many cases, the tokens are written in different forms to save space, capitalized, mixed with numbers etc. We apply the following text preprocessing steps to normalize the tweet text: 1) remove hashtags, URLs, user tags, retweets tags using Ekphrasis, 2) remove punctuations, 3) convert tokens into lowercase.

\subsection{Model Parameters}

In this section, we provide details about all parameters of the buildings blocks in the model architecture show in Figure~\ref{fig:model}.

\textbf{BERT Encoder}: We experiment with two different variants of BERT models. The first variant is \textit{BERT-base}, which is the default model provided by \citet{bert}. The second variant is \textit{HateBERT} provided by \citet{hatebert}, which is a \textit{BERT-base} model pre-trained further on hateful comments corpus extracted from Reddit. Both variants output a sequence of 768-dimensional vectors for the given input tokens.

\textbf{Recurrent Neural Network (RNN) Layers}: We experimented with different types of RNN layers: Long-short Term Memory (LSTM), Gated Recurrent Unit (GRU), and Bidirectional Gated Recurrent Unit (Bi-GRU). We also experimented with different layer sizes, which are 100, 200, 300.

\textbf{Linear Layers}: The model architecture includes four blocks that are composed of three consecutive layers: 
linear layer, batch normalization, and activation function (ReLU). The sizes of the linear layers in the \textit{Block-1}, \textit{Block-2}, \textit{Block-3}, \textit{Block-4} are 512, 512, 256, 128 respectively.

\textbf{Training Process}: Each configuration of the model architecture is trained using Adam optimizer~\cite{DBLP:journals/corr/KingmaB14} with a learning rate of \textit{0.001}, a batch size of \textit{64} for maximum of \textit{20} iterations. We use the 90:10 splits for train and validation splits to find the optimal hyperparameters.

\textbf{Implementation}: The model architecture is implemented in Python using the Tensorflow Keras library. The source code is shared openly with the community\footnote{\url{https://github.com/sherzod-hakimov/HASOC-2021---Hate-Speech-Detection}}.

\subsection{Results}

We tested different model configurations as explained above for all three datasets. The results are given in Table~\ref{tab:results}. The official evaluation metrics for the 
\textit{English Subtask 1A}~\cite{hasoc2021overview} is Macro F1-score. Additionally, we included accuracy and weighted F1-score since the number of data points for each class are not balanced for test splits of the datasets (see Table~\ref{tab:dataset_stat}). We included the best performing models with the corresponding features. Based on the initial experiments, the choice of Gated Recurrent Unit (GRU) with the layer size of \textit{100} yielded the highest performance on the validation set for three datasets in comparison to other model configurations. Therefore, all model configurations listed in the table below use the \textit{GRU} layer with the size \textit{100}.

\begin{table}[!ht]
\caption{Evaluation results of various model configurations on \textit{English Subtask 1A} of three datasets. The evaluation metrics are accuracy (Acc), Macro F1-score (M-F1), and weighted F1-score (W-F1). The best performing model configurations for each dataset are highlighted in bold. \textbf{BB}: word embeddings extracted from a pre-trained \textit{BERT-base}~\cite{bert} model, \textbf{HB}: word embeddings extracted from a pre-trained \textit{HateBERT}~\cite{hatebert} model, \textbf{CH}: character level embeddings, \textbf{HW}: multi-hot encoding of hate words.}
\begin{tabular}{l|c|l|l|c|l|l|c|l|l|}
\cline{2-10}
 & \multicolumn{3}{c|}{\textbf{HASOC-2019}} & \multicolumn{3}{c|}{\textbf{HASOC-2020}} & \multicolumn{3}{c|}{\textbf{HASOC-2021}} \\ \hline
\multicolumn{1}{|l|}{\textbf{Features}} & \textbf{Acc} & \multicolumn{1}{c|}{\textbf{M-F1}} & \multicolumn{1}{c|}{\textbf{W-F1}} & \textbf{Acc} & \multicolumn{1}{c|}{\textbf{M-F1}} & \multicolumn{1}{c|}{\textbf{W-F1}} & \textbf{Acc} & \multicolumn{1}{c|}{\textbf{M-F1}} & \multicolumn{1}{c|}{\textbf{W-F1}} \\ \hline
\multicolumn{1}{|l|}{$BB$} & \multicolumn{1}{l|}{\textbf{0.78}} & \textbf{0.70} & \textbf{0.78} & \multicolumn{1}{l|}{0.86} & 0.86 & 0.86 & \multicolumn{1}{l|}{0.74} & 0.73 & 0.73 \\ \hline
\multicolumn{1}{|l|}{$HB$} & \multicolumn{1}{l|}{0.64} & 0.61 & 0.61 & \multicolumn{1}{l|}{0.84} & 0.84 & 0.84 & \multicolumn{1}{l|}{0.74} & 0.73 & 0.74 \\ \hline
\multicolumn{1}{|l|}{$CH$} & \multicolumn{1}{l|}{0.36} & 0.36 & 0.36  & \multicolumn{1}{l|}{0.56} & 0.51 & 0.60 & \multicolumn{1}{l|}{0.63} & 0.59  & 0.64 \\ \hline
\multicolumn{1}{|l|}{$HW$} & \multicolumn{1}{l|}{0.75} & 0.61 & 0.77 & \multicolumn{1}{l|}{\textbf{0.89}} & \textbf{0.89} & \textbf{0.89} & \multicolumn{1}{l|}{0.71} & 0.71  & 0.71 \\ \hline
\multicolumn{1}{|l|}{$CH$ + $HW$} & \multicolumn{1}{l|}{0.76} & 0.59 & \underline{0.80} & \multicolumn{1}{l|}{\textbf{0.89}} & \textbf{0.89} & \textbf{0.89} & \multicolumn{1}{l|}{0.71} & 0.71 & 0.71 \\ \hline
\multicolumn{1}{|l|}{$BB$ + $HW$} & \multicolumn{1}{l|}{0.78} & 0.68 & \underline{0.80} & \multicolumn{1}{l|}{0.88} & 0.88 & 0.88 & \multicolumn{1}{l|}{0.78} &0.76 & 0.78 \\ \hline
\multicolumn{1}{|l|}{$BB$ + $CH$} & \multicolumn{1}{l|}{0.77} & 0.70 & 0.77 & \multicolumn{1}{l|}{0.81} & 0.81 & 0.81 & \multicolumn{1}{l|}{0.73} & 0.72 & 0.73 \\ \hline
\multicolumn{1}{|l|}{$HB$ + $HW$} & \multicolumn{1}{l|}{0.71} & 0.65 & 0.70 & \multicolumn{1}{l|}{\textbf{0.89}} & \textbf{0.89} & \textbf{0.89} & \multicolumn{1}{l|}{0.78} & 0.77 & 0.78 \\ \hline
\multicolumn{1}{|l|}{$HB$ + $CH$} & \multicolumn{1}{l|}{0.58} & 0.56 & 0.55 & \multicolumn{1}{l|}{0.87} & 0.87 & 0.87 & \multicolumn{1}{l|}{0.73} & 0.71 & 0.73 \\ \hline

\multicolumn{1}{|l|}{$BB$ + $CH$ + $HW$} & \multicolumn{1}{l|}{0.75} & 0.69 & 0.75 & \multicolumn{1}{l|}{\textbf{0.89}} & \textbf{0.89} & \textbf{0.89} & \multicolumn{1}{l|}{\textbf{0.79}} & \textbf{0.77} & \textbf{0.79} \\ \hline
\multicolumn{1}{|l|}{$HB$ + $CH$ + $HW$} & \multicolumn{1}{l|}{0.75} & 0.69 & 0.75 & \multicolumn{1}{l|}{0.49} & 0.33 & 0.66 & \multicolumn{1}{l|}{0.78} & 0.76 & 0.78 \\ \hline
\end{tabular}
\label{tab:results}
\end{table}

The results suggest that \textit{BERT-base} embeddings have greater a impact than \textit{HateBERT} embeddings. Another important observation is that the feature based on the multi-hot encoding of hate terms ($HW$) achieves high accuracy and weighted F1-score for all datasets. Specifically, every model configuration that included the $HW$ feature yields the best results on the \textit{HASOC-2020} dataset. Our approach that combines BERT-base, character embeddings, and multi-hot encoding of hate terms achieved a Macro F1-score of \textit{0.77} on \textit{English Subtask 1A}~\cite{hasoc2021overview} of the HASOC-2021~\cite{hasoc2021} dataset. We submitted the same model as a team \textit{TIB-VA} to the official challenge. Our model was ranked at the position \textit{33} with the Macro-F1 score of \textit{0.76}.

\begin{figure*}[ht]
\begin{subfigure}{.5\textwidth}
  \centering
  \includegraphics[width=.98\linewidth]{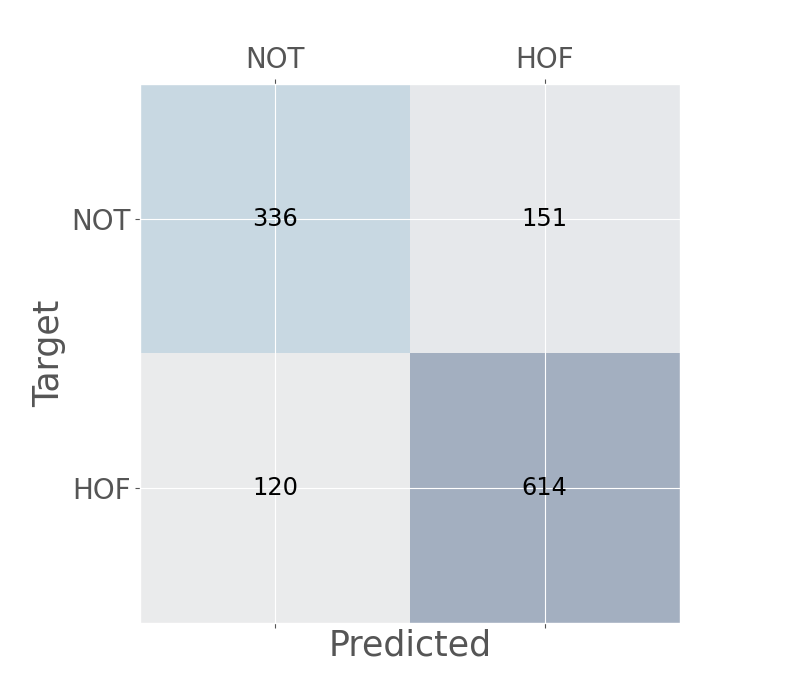}
  \caption{Model with features: $BB$ + $CH$ + $HW$}
  \label{fig:sfig1}
\end{subfigure}%
\begin{subfigure}{.5\textwidth}
  \centering
  \includegraphics[width=.98\linewidth]{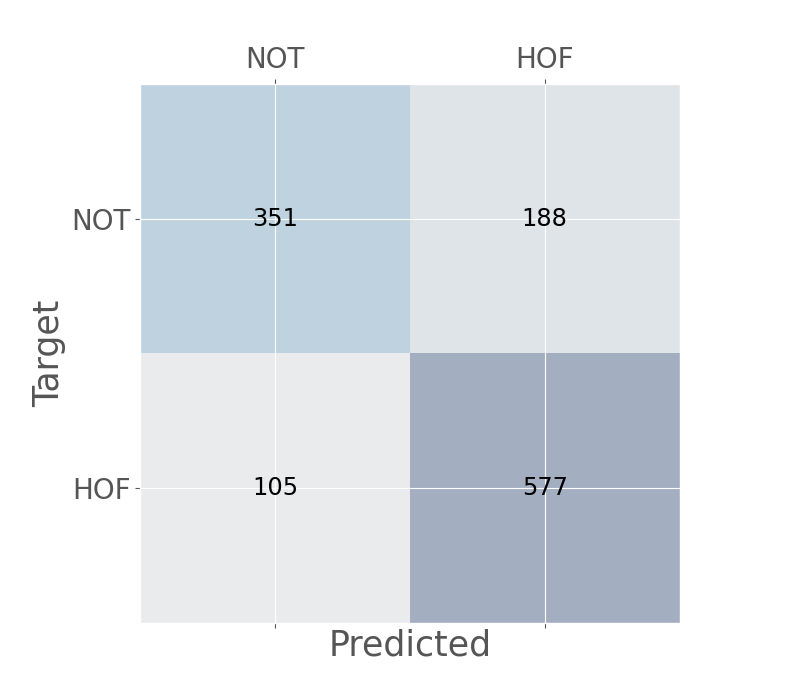}
  \caption{Model with features: $HB$ + $CH$ + $HW$}
  \label{fig:sfig2}
\end{subfigure}
\caption{Confusion matrices for two models evaluated on the \textit{HASOC 2021 - English Subtask 1A}}
\label{fig:confusion_matrices}
\end{figure*}

We present the confusion matrices in Figure~\ref{fig:confusion_matrices} for two models with different pre-trained variants of BERT models: \textit{BERT-base} (BB) and \textit{HateBERT} (HB). Both models were trained with character level embeddings (CH) and multi-hot encoded hate words (HW). We can observe that the model using  \textit{BERT-base} embeddings (Figure~\ref{fig:sfig1}) makes more correct predictions (614 vs. 577) in detecting hateful content (HOF) when compared with the other model variant (Figure~\ref{fig:sfig2}). A similar pattern exists for cases where the target class is NOT, and the model predicts HOF where the model with the BB feature makes fewer mistakes (151 vs. 188) than the other model.

\section{Conclusion}\label{sec:conclusion}

In this paper, we have analyzed model architectures that combine multiple textual features to detect hateful, offensive and profane language. Our experimental results showed that simply using the multi-hot encoding of collected 1493 hate terms yields significant performance. The combination of BERT embeddings, character embeddings, and features based on hate terms achieved the best performance on the \textit{English Subtask 1A}, HASOC 2021 dataset. Another observation of the evaluation is that a variant of the BERT model trained on domain-specific (\textit{HateBERT}) text did not improve the results in comparison to the default pre-trained model variant (\textit{BERT-base}).

\bibliography{references}

\end{document}